\pdfoutput=1

\documentclass[11pt]{article}

\usepackage[final]{acl}

\usepackage{times}
\usepackage{latexsym}

\usepackage[T1]{fontenc}

\usepackage[utf8]{inputenc}

\usepackage{microtype}

\usepackage{inconsolata}

\usepackage{graphicx}

\title{SPRING Lab IITM's submission to Low Resource Indic Language Translation Shared Task}

\author{
 \textbf{Hamees Sayed},
 \textbf{Advait Joglekar},
 \textbf{Srinivasan Umesh}
\\
 SPRING Lab, \\
 Indian Institute of Technology Madras
\\
 \small{
   \href{mailto:hameessayed71@gmail.com}{hameessayed71@gmail.com}, \href{mailto:advaitjoglekar@gmail.com}{advaitjoglekar@gmail.com}, \href{mailto:umeshs@ee.iitm.ac.in}{umeshs@ee.iitm.ac.in}
 }
}

\begin{document}
\maketitle
\begin{abstract}
We develop a robust translation model for four low-resource Indic languages: Khasi, Mizo, Manipuri, and Assamese. Our approach includes a comprehensive pipeline from data collection and preprocessing to training and evaluation, leveraging data from WMT task datasets, BPCC, PMIndia, and OpenLanguageData. To address the scarcity of bilingual data, we use back-translation techniques on monolingual datasets for Mizo and Khasi, significantly expanding our training corpus. 
We fine-tune the pre-trained NLLB 3.3B model for Assamese, Mizo, and Manipuri, achieving improved performance over the baseline. For Khasi, which is not supported by the NLLB model, we introduce special tokens and train the model on our Khasi corpus. Our training involves masked language modelling, followed by fine-tuning for English-to-Indic and Indic-to-English translations.
\end{abstract}

\section{Introduction}

Translation of low-resource languages poses significant challenges in natural language processing. While substantial progress has been made in developing machine translation models for high-resource languages, low-resource languages often suffer from a lack of parallel corpora and digital resources~\cite{haddow-etal-2022-survey}. Languages like Khasi, Mizo, Manipuri, and Assamese are representative of this challenge, where limited data and unique linguistic complexities hinder the development of robust translation systems.

In recent years, efforts to bridge this gap have gained momentum, driven by initiatives such as the Bharat Parallel Corpus Collection\footnote{\url{https://ai4bharat.iitm.ac.in/bpcc/}} (BPCC) ~\cite{gala2023indictrans2} and government-supported projects like PMIndia~\cite{haddow2020pmindiacollectionparallel}, which aim to provide bilingual data for Indic languages. Despite these efforts, translation models for low-resource Indic languages have yet to achieve performance levels comparable to their high-resource counterparts~\cite{suman-etal-2023-iacs}, necessitating innovative approaches to model training and data utilization.

In this work, we develop a robust translation model for four low-resource Indic languages: Khasi, Mizo, Manipuri, and Assamese. Our approach involves data collection, preprocessing, training, and evaluation. We utilize datasets from WMT, BPCC, PMIndia, and OpenLanguageData\footnote{\url{https://github.com/openlanguagedata/seed}}~\cite{seed-23}, and enhance bilingual data through back-translation~\cite{edunov2018understandingbacktranslationscale} techniques, especially for Mizo and Khasi, significantly expanding our training corpus.

We follow Meta's data preprocessing standards and use LoRA (Low-Rank Adaptation) ~\cite{hu2021loralowrankadaptationlarge} fine-tuning on the NLLB~\cite{nllbteam2022languageleftbehindscaling} 3.3B model to improve efficiency and performance with fewer parameters. Our model initially focuses on one-way translation from English to the Indic languages, then on reverse translations~\cite{dabre-etal-2019-exploiting}. The results show improved performance over the baseline, particularly for Khasi, where we address gaps in pre-trained model support.

\section{Dataset}

In this study, we focus on four low-resource Indic languages covered in the Low Resource Indic Languages Shared Task: Khasi, Mizo, Manipuri, and Assamese. This section highlights the significance of each language, including their role in their respective regions, their linguistic and cultural importance, and the details of the datasets used. Statistics regarding language speakers are according to the 2011 Indian Census\footnote{\url{https://censusindia.gov.in/}}.

\begin{table*}
    \centering
    \begin{tabular}{lccccccc}
         \textbf{Language} & \textbf{ISO-693-3} & \textbf{WMT Parallel} & \textbf{BPCC} & \textbf{PMIndia} & \textbf{OLD} & \textbf{Back-Translated} & \textbf{Total} \\
         \hline
         Assamese  & asm & 50,000 & 35,354 & 9,732 & 0     & 0 & 95,086 \\
         Manipuri  & mni & 21,687 & 0      & 7,419 & 6,193 & 0 & 35,036 \\
         Khasi     & kha & 24,000 & 0      & 0     & 0     & 102,070 & 126,070 \\
         Mizo      & lus & 50,000 & 0      & 0     & 0     & 30,164 & 80,164 \\
         \hline
    \end{tabular}
    \caption{Breakdown of data sources and volumes for each language. ``OLD'' refers to OpenLanguageData. The ``Back-Translated'' data was initially generated using Google Translate\protect\footnotemark{} for the first 500k characters from the monolingual WMT task data, with subsequent iterations increasing the data size using the trained model}
    \label{tab:data}
\end{table*}

\subsection{Languages}

\textbf{Assamese} \textit{(Asamiya)} is an Indo-Aryan language spoken primarily in the northeastern Indian state of Assam, where it serves as an official language and a regional lingua franca. With over 15 million native speakers, it is one of the most widely spoken languages in the region. Historically, Assamese was the court language of the Ahom kingdom. It is written in the Assamese script, an abugida system, known for its unique typographic ligatures. \\

\noindent
\textbf{Manipuri} \textit{(Meiteilon)} is a key Tibeto-Burman language spoken mainly in Manipur, India, where it is an official language and it is one of the constitutionally scheduled official languages of the Indian Republic. With 1.76 million speakers, it is the most widely spoken Tibeto-Burman language in India and holds the third place among the fastest-growing languages of India, following Hindi and Kashmiri. It is written in its own Meitei script as well as the Bengali script.  \\

\noindent
\textbf{Khasi} \textit{ (Ka Ktien Khasi)} is an Austroasiatic language primarily spoken by the Khasi people in Meghalaya, India, with approximately 1 million native speakers as of the 2011 census. The language holds an associate official status in certain districts of Meghalaya. Khasi is written in the Latin script. The closest relatives of Khasi are other languages in the Khasic group, such as Pnar and War.  \\  

\noindent
\textbf{Mizo} \textit{(Mizo ṭawng)} belonging to the Sino-Tibetan language family, is primarily spoken in the state of Mizoram, India, with around 800 thousand speakers. The Mizo language, also known as Lushai, has a rich oral history and was first written using the Latin script in the late 19th century. Mizo is recognized as the official language of Mizoram and is used in education, government, and media.  \\

\footnotetext{\protect\url{https://google.translate.com/}}
\section{Methodology}

This section covers the preprocessing steps and training methods used, including dataset preparation and the fine-tuning of Meta's multilingual NLLB 3.3B base pre-trained model. Detailed statistics on data distribution are presented in Table \ref{tab:data}.

\subsection{Preprocessing}

In the preprocessing phase, we followed a series of steps to ensure the text data was clean and consistent before model training. We began by normalizing punctuation using Moses~\cite{koehn-etal-2007-moses}, an open-source toolkit designed for preprocessing, training, and testing translation models. This step helps maintain consistency in text data, which is crucial for training robust models.

Non-printable characters, which often interfere with text processing, were replaced with a space. This choice ensures that any invisible or non-standard characters do not disrupt the tokenization process and ensures that the text is composed of standard printable characters.

We also applied Unicode normalization (NFKC) to transform characters into their canonical forms, making the text more uniform across different Unicode representations.

These preprocessing steps are aligned with those outlined by Meta for their multilingual models, and further details can be found on their GitHub\footnote{\url{https://github.com/facebookresearch/stopes/blob/main/stopes/pipelines/monolingual/monolingual_line_processor.py}}. This approach ensures that the text data used for training is clean, consistent, and compatible with the modelling requirements.

\subsection{Training} 

For model training, we employed Meta's NLLB (No Language Left Behind) 3.3B parameter model, a state-of-the-art multilingual machine translation model built to support over 200 languages, making it ideal for tasks involving low-resource languages~\cite{tran2021facebookaiwmt21news, yang-etal-2021-multilingual-machine}. The NLLB 3.3B model is based on a Transformer~\cite{vaswani2023attentionneed} architecture with 3.3 billion parameters, featuring a dense encoder-decoder design. It includes the following hyperparameters:

\begin{table}[h]
  \centering
  \begin{tabular}{lc}
    \textbf{Hyperparams} & \textbf{} \\
    \hline
    embed size     & 2048           \\
    ffn size     & 8192           \\
    attn heads     & 16           \\
    enc/dec layers     & 48           \\
    \hline
  \end{tabular}
  \caption{Hyperparameters for the baseline pre-trained model. 24 Encoder and 24 Decoder Layers.}
  \label{tab:hyperparams}
\end{table}

To fine-tune the model, we employed LoRA, a technique that significantly reduces computational demands and training time by adapting only a small subset of the model's parameters. LoRA has been shown to match the performance of traditional fine-tuning methods while reducing the number of trainable parameters by a factor of 50~\cite{alves2023steeringlargelanguagemodels}. This approach is especially effective for large-scale models like Meta's NLLB 3.3B, allowing efficient adaptation without significantly compromising on performance.

\subsection{Parameters}

The training process was conducted in three stages: first, the model was trained on masked language modelling~\cite{devlin2019bertpretrainingdeepbidirectional} to enhance its understanding of the target language by leveraging monolingual data. Next, it was fine-tuned for English-to-Indic translations, followed by further fine-tuning for Indic-to-English translations. In the case of Khasi, which was not natively supported by the NLLB model, special tokens were added to the tokenizer's vocabulary to accommodate the Khasi language. The model was subsequently trained on the Khasi corpus to ensure proper handling and integration of this language.

\begin{table}[ht]
  \centering
  \begin{tabular}{lc}
    \textbf{Training Args} & \textbf{} \\
    \hline
    optimizer     & adafactor           \\
    learning Rate     & 1e-5           \\
    epochs     & 8           \\
    precision     & bf16           \\
    $p_{mask}$   &  0.15 \\
    peft type   &  lora \\
    rank    & 128         \\
    lora alpha    & 256        \\
    lora dropout    & 0.1           \\
    target modules    & all linear   \\
    \hline
  \end{tabular}
  \caption{Training parameters and LoRA configuration used for fine-tuning the NLLB 3.3B model.}
  \label{tab:trainingargs}
\end{table}

The training was performed across 4 Nvidia A6000 GPUs. These settings allowed us to optimize the model's performance while managing computational efficiency.

\subsection{Inference}

For inference, the trained adapter was loaded onto the NLLB 3.3B model. The model generated predictions using a beam search strategy with 10 beams and a repetition penalty of 2.5 to improve the diversity and quality of the translations. We experimented with various beam and penalty configurations, ultimately finding that this particular setup produced the most accurate and linguistically coherent outputs.

\section{Results}

The evaluation of our translation model across various language pairs and directions is shown in Table \ref{tab:results}, with performance assessed using BLEU \cite{papineni-etal-2002-bleu}, Translation Error Rate \cite{snover-etal-2006-study}, RIBES \cite{isozaki-etal-2010-automatic}, METEOR \cite{banerjee-lavie-2005-meteor}, and chrF \cite{popovic-2015-chrf} metrics. We found that the scores in English-to-Manipuri and English-to-Mizo direction suffered from the poor quality of backtranslated data used in our training.

\begin{table*}[t]
\centering
\begin{tabular}{lcccccc}
\textbf{Language Pairs} & \textbf{Test Set} & \textbf{BLEU} & \textbf{TER} & \textbf{RIBES} & \textbf{METEOR} & \textbf{ChrF} \\
\hline
{English-Assamese} & en\_to\_as\_contrastive & 27.26 & 52.79 & 0.3032 & 0.513 & 65.2 \\
 & as\_to\_en\_contrastive & 26.69 & 39.08 & 0.3308 & 0.7066 & 60.48 \\
\hline
 {English-Manipuri} & en\_to\_mn\_contrastive & 2.7 & 84.6 & 0.1185 & 0.1567 & 44.28 \\
 & mn\_to\_en\_contrastive & 20.88 & 48.77 & 0.3031 & 0.61 & 53.64 \\
 \hline
{English-Khasi} & en\_to\_kh\_contrastive & 12.12 & 63.31 & 0.1864 & 0.4453 & 44.55 \\
 & kh\_to\_en\_contrastive & 10.47 & 61.43 & 0.2172 & 0.5042 & 42.71 \\
 \hline
{English-Mizo} & en\_to\_mz\_contrastive & 6.6 & 66.06 & 0.1746 & 0.495 & 49.79 \\
 & mz\_to\_en\_contrastive & 18.49 & 53.19 & 0.2684 & 0.588 & 50.44 \\
\hline
\end{tabular}
\caption{Translation performance metrics of our MT System reported in the final evaluation.}
\label{tab:results}
\end{table*}

\noindent
\textbf{English-Assamese} The model performed relatively well, with BLEU scores of 27.26 for English-to-Assamese and 26.69 for Assamese-to-English. \\   
\textbf{English-Manipuri} The model showed lower BLEU scores for English-to-Manipuri (2.7) compared to Manipuri-to-English (20.88). The TER score was higher for English-to-Manipuri, reflecting greater translation errors in this direction.   \\   
\textbf{English-Khasi} For Khasi, the BLEU score was 12.12 for English-to-Khasi and 10.47 for Khasi-to-English.\\   
\textbf{English-Mizo} The performance was mixed, with a BLEU score of 6.6 for English-to-Mizo and 18.49 for Mizo-to-English. The TER score indicates a higher error rate for English-to-Mizo, while the METEOR and ChrF scores were relatively balanced across both directions.   

\section{Conclusion}

In this work, we utilized Meta's NLLB 3.3B model, a large-scale multilingual transformer with 3.3 billion parameters, to enhance translation between low-resource Indic languages and English. The training process included masked language modelling, followed by English-to-Indic and Indic-to-English translations. Special tokens were added for Khasi, and LoRA (Low-Rank Adaptation) was employed to optimize computational efficiency and reduce training time. \\
Conducted on 4 NVIDIA A6000 GPUs, our approach demonstrates that large-scale multilingual models, when combined with LoRA, effectively capture diverse linguistic patterns and advance translation capabilities.

\section{Limitations}

In this study, we encountered several limitations that impacted the overall effectiveness of our translation system. One major challenge was the constrained size of our dataset due to computational resource limitations. The limited dataset size may have hindered the model's ability to generalize, particularly for low-resource languages where larger and more diverse datasets would have been advantageous.

Another issue we faced was the quality of backtranslated data. The process of augmenting the dataset through machine translation often resulted in lower-quality data, which negatively influenced the model's performance. This highlights the need for more robust data generation techniques in future work.

We also observed a noticeable performance gap between translations where English was the target language and those where an Indic language was the target. This suggests that while the model may understand the morphological aspects of Indic languages, it struggles to generate accurate translations in these languages. This limitation underscores the need for further refinement in handling the complexities of Indic language generation.

Finally, the potential domain mismatch between our training data and real-world applications posed a significant challenge. The training data may not fully capture the linguistic and contextual nuances found in practical scenarios, leading to reduced performance in actual use cases. Addressing this issue in future work will be crucial for improving the model's real-world applicability.

\nocite{pal2023findings}
\nocite{pakray2024findings}
\bibliography{anthology, latex/custom}

\end{document}